\documentclass[lettersize,journal]{IEEEtran}
\usepackage{amsmath,amsfonts}
\usepackage{algorithmic}
\usepackage{algorithm}
\usepackage{array}
\usepackage[caption=false,font=normalsize,labelfont=sf,textfont=sf]{subfig}
\usepackage{textcomp}
\usepackage{stfloats}
\usepackage{url}
\usepackage{verbatim}
\usepackage{graphicx}
\usepackage{booktabs}
\usepackage{multirow}
\usepackage[mathscr]{eucal}
\usepackage{cite}
\usepackage{tcolorbox}
\usepackage{xcolor}
\usepackage{hyperref}
\hypersetup{hidelinks,
	colorlinks=true,
	allcolors=black,
	pdfstartview=Fit,
	breaklinks=true}

\begin{document}

\title{TAMO:Fine-Grained Root Cause Analysis via Tool-Assisted LLM Agent with Multi-Modality Observation Data in Cloud-Native Systems
}



\author{Xiao Zhang, \and Qi Wang,  \and Mingyi Li, \and Yuan Yuan, \and Mengbai Xiao,  \and Fuzhen Zhuang, \and and Dongxiao Yu 

\thanks{
This work was supported in part by the National Natural Science Foundation of China under Grant 62202273, 62176014,  in part by the Joint Key Funds of National Natural Science Foundation of China under Grant U24B20149, in part by Major Basic Research Program of Shandong Provincial Natural Science Foundation under Grant ZR2025ZD18, in part by China Postdoctoral Science Foundation under Grant 2024M761806, and in part by the Fundamental Research Funds for the Central Universities.  (\textit{Corresponding author: Dongxiao Yu.})
}

\thanks{Xiao Zhang, Qi Wang, Mingyi Li, Mengbai Xiao, and Dongxiao Yu are with the School of Computer Science and Technology, Shandong University, Qingdao 266237, China (Email:
xiaozhang@sdu.edu.cn; wsq5457@mail.sdu.edu.cn; limee@mail.sdu.edu.cn; xiaomb@sdu.edu.cn; dxyu@sdu.edu.cn).}
\thanks{Yuan Yuan is with the School of Software $\&$ Joint SDU-NTU Centre for Artificial Intelligence Research (C-FAIR), Shandong University, Jinan, 250000, China (Email: yyuan@sdu.edu.cn).}
\thanks{
Fuzhen Zhuang is with the Institute of Artificial Intelligence, Beihang University, Beijing, 100191, China, and State Key Laboratory of Complex \& Critical Software Environment (SKLCCSE), Beihang University, Beijing 100191, China.
(Email: zhuangfuzhen@buaa.edu.cn).}
}

\markboth{Journal of \LaTeX\ Class Files,~Vol.~14, No.~8, August~2021}%
{Shell \MakeLowercase{\textit{et al.}}: A Sample Article Using IEEEtran.cls for IEEE Journals}

\IEEEpubid{0000--0000/00\$00.00~\copyright~2021 IEEE}

\maketitle

\begin{abstract}
Implementing large language models (LLMs)-driven root cause analysis (RCA) in cloud-native systems 
has become a key topic of modern software operations and maintenance. 
However, existing LLM-based approaches face three key challenges: 
multi-modality input constraint, context window limitation, and dynamic dependence graph. 
To address these issues, we propose a tool-assisted LLM agent with multi-modality observation data for fine-grained RCA, namely TAMO, including multi-modality alignment tool, root cause localization
tool, and fault types classification tool.  
In detail, TAMO unifies multi-modal observation data into time-aligned representations for cross-modal feature consistency.  
Based on the unified representations, TAMO then invokes its specialized root cause localization tool  and fault types classification tool for further identifying root cause and fault type underlying system context.
This approach overcomes the limitations of LLMs in processing real-time raw observational data and dynamic service dependencies, guiding the model to generate repair strategies that align with system context through structured prompt design.
Experiments on two benchmark datasets demonstrate that TAMO outperforms state-of-the-art (SOTA)
approaches with comparable performance.
\end{abstract}


\section{Introduction}
\IEEEPARstart{I}{n} recent years, 
microservice architecture in cloud-native systems has become foundational elements in modern enterprise software development, driving the rapid 
improvement of distributed systems 
\cite{li2024unity}\cite{luo2021characterizing}\cite{soldani2018pains}\cite{YU2023100150}. Microservices decompose traditional monolithic applications into independent distributed services, while cloud native technologies leverage containerization and container orchestration to enhance the capabilities for flexible resource allocation and autoscaling \cite{khan2017key}\cite{NOOR2025100276}. 
In detail, cloud-native systems are composed of numerous independent microservices and containerized entities (e.g., pod, node)\cite{zhou2018fault}, which inevitably 
brings inter-service dependencies and complex communication patterns. 
Therefore, the failure of any entity (e.g., network corruption) can trigger cascading effects that damage overall system stability, potentially leading to significant economic losses\cite{amin2013approach}\cite{SONG2025100268}. 



\begin{figure}[!t]
\centering
\includegraphics[width=0.48\textwidth]{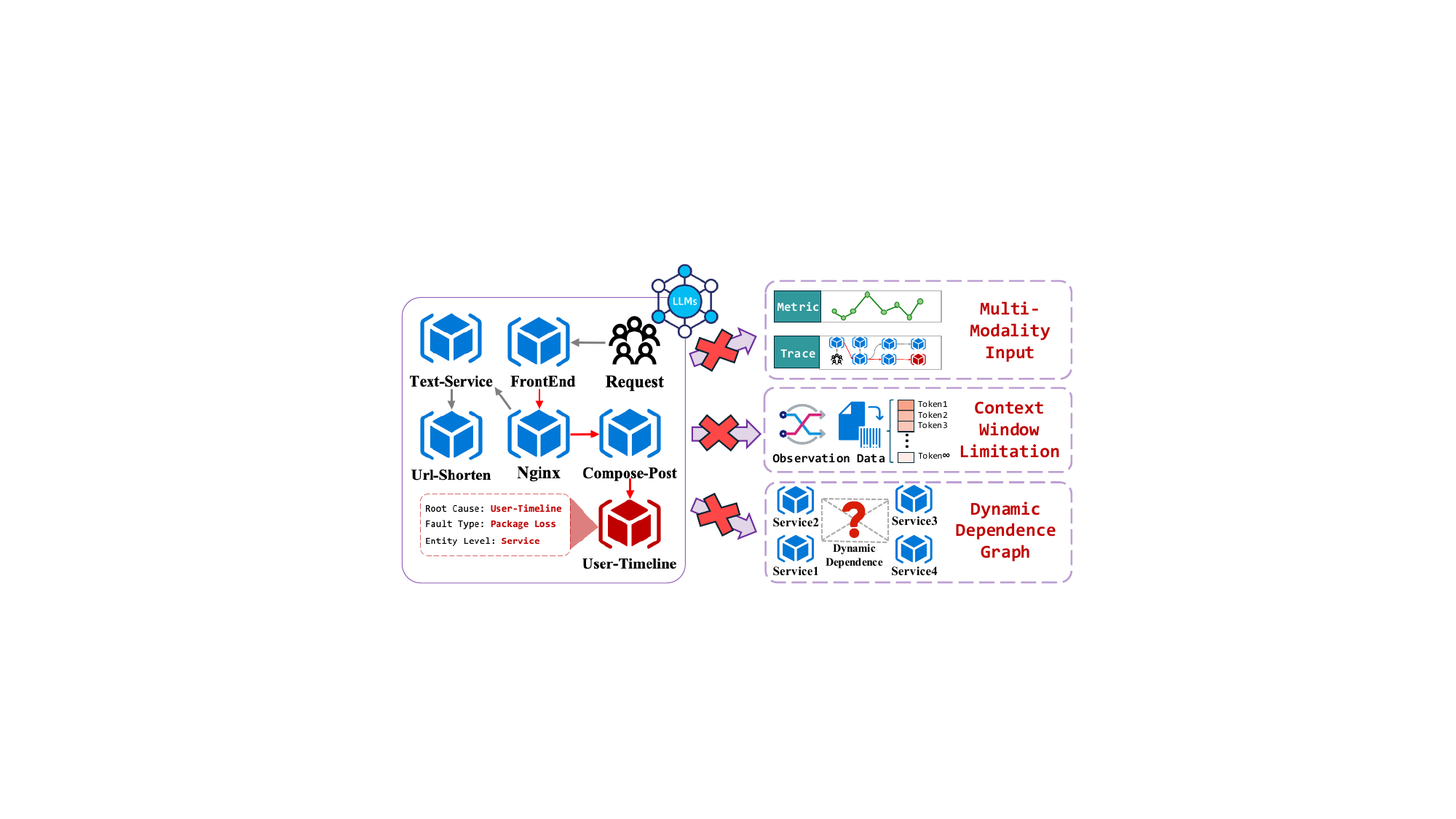}
\caption{A failure case in a cloud-native system demonstrates three major challenges when using LLM directly for root cause analysis.}
\label{fig:intro}
\end{figure}

\IEEEpubidadjcol
Root cause analysis (RCA) aims to quickly correlate features of multiple related entities, localize root cause, identify corresponding fault type, and provide reasonable 
solutions, thus realizing an automatic fault response mechanism and minimizing human intervention \cite{CHEN2022100050}\cite{zhang2023robust}. 
Therefore, how to perform accurate and efficient root cause analysis in 
cloud-native systems 
has become an important research topic recently \cite{han2024holistic}\cite{zheng2024mulan}.  
On one hand, traditional RCA methods \cite{liu2021microhecl} \cite{marwede2009automatic} typically rely on rule-based anomaly detection or statistical correlation analysis (e.g. causal graphs)\cite{li2024realtcd} to localize faults. These methods struggle to jointly analyze heterogeneous data modalities with unstructured logs and 
metric data, often leading to undetected faluts. 
On the other hand, deep learning (DL) based RCA methods attempt to incorporate multimodal feature fusion and dependency-sensitive graph neural networks (GNNs)\cite{lee2023eadro}\cite{zhang2024rethinking} to achieve accurate fault localization.
For instance, MULAN~\cite{zheng2024mulan} infers causal structures from heterogeneous observation data for service-level diagnosis, while PDiagnose~\cite{hou2021diagnosing} treats multimodal data independently and aggregates decisions via voting to pinpoint root causes. 
The abovementioned methods mainly focus on specific tasks such as fault localization in complex systems, ignoring fine-grained RCA with fault type identification, contextual anomaly interpretation, and cross-level dependency reasoning. 
Recently, large language models (LLMs) have demonstrated revolutionary capabilities in automated RCA, opening new avenues for AIOps \cite{cheng2023ai}\cite{fan2023large}. LLMs can combine RCA 
with current system context to generate reasonable fault diagnosis reports, emerging with RCACopilot\cite{chen2023automaticrootcauseanalysis}. 
Recent works such as RCAgent\cite{wang2024rcagent} and mABC\cite{zhang2024mabc} develop LLM-driven automated RCA frameworks that enable accurate fault localization and provide reasonable repair suggestions, thus further advancing the field of automated operations and maintenance. RCAgent proposes a tool-augmented autonomous agent framework that leverages log and code data with an internally deployed LLM to perform root cause analysis, while mABC mitigates circular dependency issue in microservice RCA through blockchain-inspired multi-agent voting within a structured workflow. 

However, existing LLM-based methods still face some tricky challenges when applied to fine-grained RCA in cloud-native systems involving multi-level entities and heterogeneous multi-modality data: 
\begin{itemize}    
    \item \textbf{Multi-Modality Input}: 
    Existing LLM-driven methods such as COCA\cite{li2025coca}, RCAgent\cite{wang2024rcagent},
    only support log data or code data as input, ignoring abundant context information such as metric data and trace data. 
    However, directly feeding multimodal data into LLMs often hinders accurate RCA due to inherent modality misalignment, for instance, the semantic gap between time-series metrics and textual logs.
    Therefore, 
    \textit{how to map different modalities into unified semantic space  to assist the LLM in accurate RCA} is challenging. 

    \item \textbf{Context Window Limitation}: Limited by fixed-length context windows, the LLM is unable to effectively process the massive real-time metric data generated by continuous monitoring across kinds of microservices, nodes or pods  
    in cloud-native systems. 
    The real-time metric data  typically reflect system states but does not explicitly indicate anomalies or fault types, 
    yet lacking aligned semantic meaning with text logs.
    \textit{So how to transform raw measurements into meaningful diagnostic prompts and input them  into the LLM with limited context length?} 
    \item \textbf{Dynamic Dependence Graph}:
    The dynamic service call chains from the trace data in the system inevitably bring evolving dependencies. 
    However, exsiting LLM-based methods \cite{wang2023can} struggle with fine-grained RCA over dynamic dependence graphs, making it difficult to capture implicit fault propagation path. 
    \textit{So how to assist the LLM agent in modeling fault propagation paths based on dynamic dependence graph to improve 
    root cause localization and fault types classification?}
\end{itemize}

To address these chanllenges, we propose a tool-assisted LLM agent framework based on kinds of domain-specific tools. 
This framework decouples the LLM from the raw multi-modality observation data by utilizing specialized
tools: multi-modality alignment tool $\mathscr{T}_1$, root cause localization
tool $\mathscr{T}_2$, and fault types classification 
tool $\mathscr{T}_3$. 
These specialized model tools are used to perceive the contextual environment, and their results form structured prompts 
that are fed back to the RCA expert agent for further analysis, generating accurate root cause analysis results and reasonable maintenance recommendations. 
In detail, multi-modality alignment tool $\mathscr{T}_1$ is based  on a 
dual-branch diffusion architecture, 
which unifies multi-modality observation data, such as log data, trace data and metric data, into temporally consistent representations. 
To address dynamic dependencies challenge, the root cause localization tool ($\mathscr{T}_2$) jointly utilizes the extracted representations from $\mathscr{T}_1$ 
and dependence graphs extracted from trace data 
to model fault propagation and identify the root cause via frequency-domain attention-driven temporal causal analysis.
Subsequently, the fault types classification tool ($\mathscr{T}_3$) further utilizes the casual graph from $\mathscr{T}_2$ to identify the anomalous patterns. 

The main contributions are summarized as follows: 
\begin{itemize} 
    \item We propose a domain-specific tools assisted LLM agent framework for fine-grained RCA, including multi-modality alignment tool, root cause localization
     tool, and fault types classification tool to empower the LLM for real-time system diagnosis.
    \item In detail, we present a dual-branch collaborative diffusion tool for multi-modality alignment that captures potentially consistent patterns across modalities and provides unified representations for downstream tasks through condition-oriented collaborative reconstruction. Otherwise, the root cause localization tool is based on frequency-domain attention mechanisms, which identifies causal dependency patterns in frequency-domain features while filtering out low-frequency noise, achieving precise root cause localization. 
    \item Experiments on two benchmark datasets demonstrate that our proposed TAMO outperforms 
    state-of-the-art (SOTA) approaches,  
    achieving average \textit{Acc@1} improvement of $4.8\%$ for root cause localization and \textit{MiPr} improvement of $10.8\%$ for fault types classification respectively. 
\end{itemize}

\section{Related Work}
In modern cloud-native systems, with the widespread adoption of microservice architectures and dynamic infrastructures, it has become particularly important to ensure high system availability and rapid fault recovery, making root cause analysis (RCA) a critical part of operations and maintenance. As a result, various RCA methods have been developed to guarantee system reliability, including traditional RCA methods, multi-modal RCA methods, and LLM-based RCA methods.

\textbf{Traditional RCA methods}: Traditional RCA methods are mainly based on statistical correlation analysis techniques, such as anomaly detection based on predefined rules and root cause localization strategies based on causal discovery\cite{chen2019outage}\cite{yang2019microservices}. While these approaches perform well in static environments, they have difficulties in adapting to the dynamic service dependencies that characterize cloud native environments. 
With the rapid advancement of deep learning, neural network-based methods for real-time system data analysis continue to emerge. DeepLog\cite{du2017deeplog} models log features via long short-term memory networks. GDN\cite{deng2021graph} and GTA\cite{chen2021learning} construct graph neural networks using metrics data to capture fault propagation paths. Sage\cite{gan2021sage} leverages causal Bayesian networks with trace data for root cause localization. However, these approaches predominantly focus on unimodal data, failing to exploit complementary inter-modal relationships.

\textbf{Multi-modal RCA methods}: The multimodal-based approach significantly reduces the undetected fault rate by comprehensively considering system observation data from different modalities. 
MULAN~\cite{zheng2024mulan} learns causal structures from logs and metrics for service-level localization. PDiagnose\cite{hou2021diagnosing} analyzes different modality data as independent entities and uses a voting mechanism to identify the root cause entity. Although this method successfully utilizes data from multiple modalities, it overlooks the consistency relationships between features from different modalities. Eadro\cite{lee2023eadro} performs gated fusion of features extracted from different modalities and uses a graph neural network to fuse embedded features at the service level for root cause localization. HolisticRCA\cite{han2024holistic} compared to Eadro, fully considers the heterogeneous characteristics of cloud native systems and standardizes the embedding of different modality data using an assembling building blocks strategy, combining masked embeddings for overall root cause analysis. In contrast to directly using raw data for root cause localization, Nezha\cite{yu2023nezha} and DiagFusion\cite{zhang2023robust} convert heterogeneous multimodal data into homogeneous event representations, performing joint analysis through event graphs to achieve root cause localization. CoE\cite{yao2024chain} further supports the operational experience of site reliability engineers as input on the event graph, enhancing the understanding and interpretability of the event. 

\textbf{LLM-based RCA methods}: Large language models are increasingly leveraged for root cause analysis  due to their strong reasoning capabilities. Recent works explore various LLM applications: Adarma~\cite{sarda2023adarma} combines LLMs with traditional ML for microservice anomaly detection and remediation using logs and metrics.
RCAcopilot~\cite{chen2023automaticrootcauseanalysis} maps multimodal observation data to predefined events and predicts root cause categories in production through an LLM-driven workflow,
and agent-based approaches~\cite{roy2024exploring} utilize tool-assisted LLMs to actively retrieve diagnostic data like logs and metrics for localization. 
Despite these advancements, significant challenges persist. The limited context window of LLMs restricts their ability to process large scale of observation data, while their inherent text-based input modality makes handling time series metrics difficult, risking loss of critical temporal patterns. Furthermore, existing methods often face limitations in generalizability (e.g., reliance on predefined rules~\cite{chen2023automaticrootcauseanalysis}),lack integration of crucial trace data for precise localization~\cite{roy2024exploring}\cite{sarda2023adarma}. These challenges highlight the need for frameworks that can effectively unify multimodal data (logs, metrics, traces) within an LLM architecture to achieve both  fine-grained root cause localization and  accurate classification. We compare these existing methods with our approach in Table~\ref{tab:comparison_vertical}.

\begin{figure*}[!t]
\centering
\includegraphics[width=0.88\textwidth]{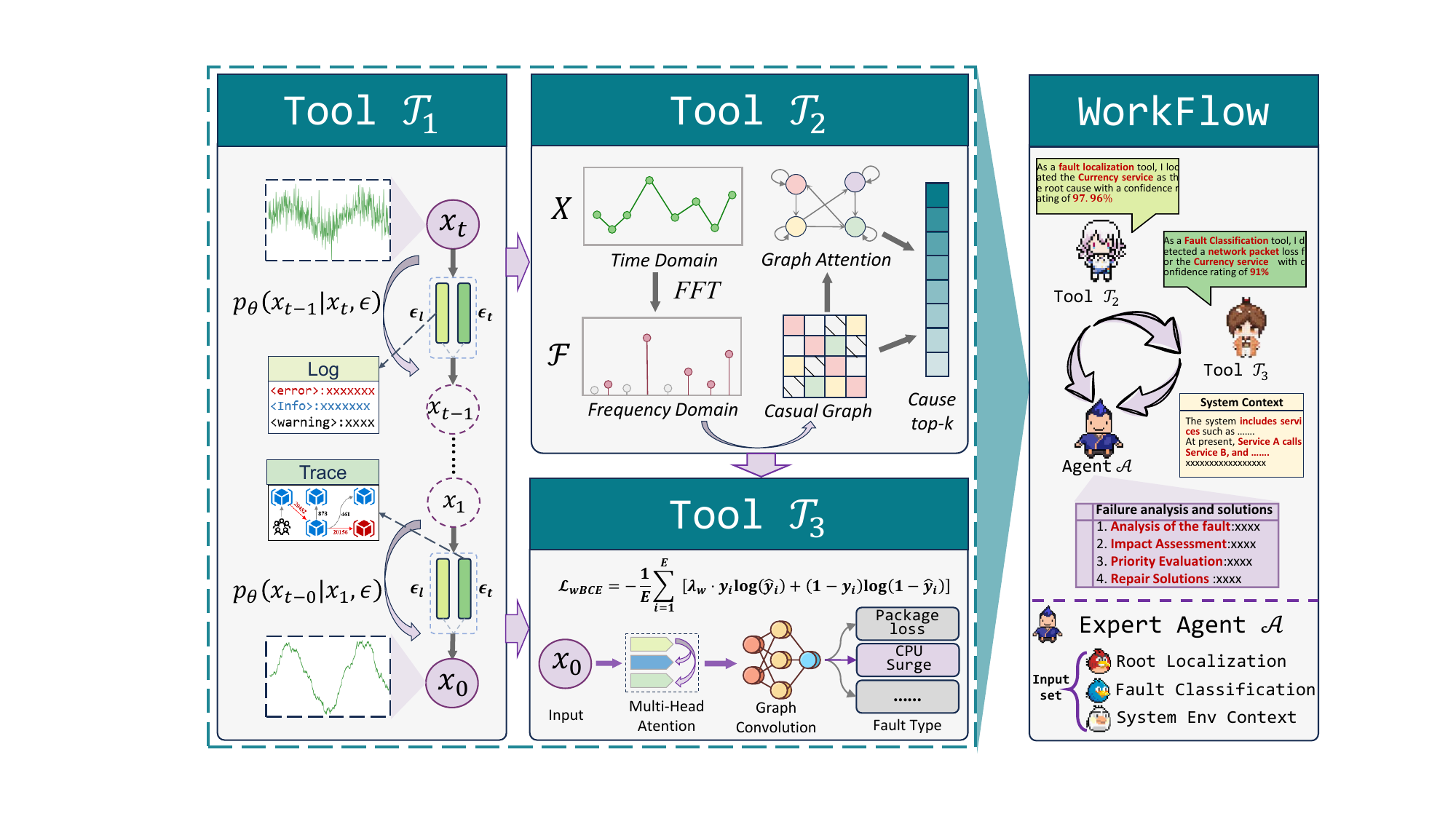}
\caption{The proposed TAMO framework consists of a Multi-modality Alignment Tool ($\mathscr{T}_1$), a Root Cause Localization Tool ($\mathscr{T}_2$), a Fault Types Classification Tool ($\mathscr{T}_3$), and an RCA Expert Agent ($\mathscr{A}$). In the framework, the agent $\mathscr{A}$ calls tools $\mathscr{T}_1$-$\mathscr{T}_3$ as perception tools to analyze the system contextual observation data in real time. The results of these perceptions are structured into text inputs that are fed back into the agent. By analyzing the perception results, the expert agent will provide the corresponding root cause analysis and repair suggestions.}
\label{fig:freamwork}
\end{figure*}

\section{METHODOLOGY}

\begin{table}[htbp]
\centering
\caption{Comparison of Different LLM RCA Models by Modality and Task. \textbf{FTC} denotes Fault Types Classification; \textbf{RCL} denotes Root Cause Localization}
\label{tab:comparison_vertical}
\begin{tabular}{lccccc}
\toprule
\multirow{3}{*}{\textbf{Methods}} &
\multicolumn{3}{c}{\textbf{Modality}} &
\multicolumn{2}{c}{\textbf{Task}} \\
\cmidrule(lr){2-4} \cmidrule(lr){5-6}
&
\textbf{Logs} & \textbf{Metrics} & \textbf{Traces} &
\textbf{FTC} & \textbf{RCL} \\
\midrule
RCAcopilot\cite{chen2023automaticrootcauseanalysis} & \checkmark & \checkmark & \checkmark & \checkmark & $\times$ \\
Agent Work\cite{roy2024exploring}               & \checkmark & \checkmark & $\times$   & $\times$   & \checkmark \\
Adarma\cite{sarda2023adarma}                & \checkmark & \checkmark & $\times$   & $\times$   & $\times$ \\
Ours                                   & \checkmark & \checkmark & \checkmark & \checkmark & \checkmark \\
\bottomrule
\end{tabular}
\end{table}
To fully exploit the multi-modality observation data of cloud native systems while integrating prior domain knowledge, we propose \textbf{TAMO}, \textbf{T}ool-Assisted LLM \textbf{A}gent that Integrates \textbf{M}ulti-Modality
\textbf{O}bservation Data to address root cause entity localization and fault types classification in 
cloud-native systems.
This framework utilizes observation tools composed of a dual-branch collaborative diffusion model to effectively extract and integrate multimodal data from different entities. Additionally, task-specialized tools explicitly model the complex dependencies among entities using a causal graph, with feature propagation through the causal graph enabling accurate root cause analysis. Furthermore, to fully incorporate domain knowledge, we employ a LLM with operational expertise as an expert agent. 
The expert agent $\mathscr{A}$ integrates the outputs from the preceding tools $\mathscr{T}_2$-$\mathscr{T}_3$ with context knowledge of the system to offer human engineers a comprehensive fault analysis and recommended solutions.
As shown in Fig.\ref{fig:freamwork}, the proposed framework consists of three tools and one agent, $\mathscr{T}_1$–$\mathscr{T}_3$ and $\mathscr{A}$, each responsible for a specific task in the root cause analysis process: 
 \begin{enumerate}
     \item Multi-modality Alignment Tool ($\mathscr{T}_1$): This tool employs a dual-branch collaborative diffusion model to integrate multimodal features. It reconstructs and fuses textual data features, such as logs, into time-series data based on conditional guidance;
     \item Root Cause Localization Tool ($\mathscr{T}_2$): Utilizing the multimodal-enhanced time-series data as input, this tool models causal relationships between entities using a frequency-domain self-attention module, enabling accurate fault root cause localization;
     \item Fault Types Classification Tool ($\mathscr{T}_3$): Based on the causal relationships between entities and the calling chain topology, this tool classifies the types of faults;
     \item RCA Expert Agent ($\mathscr{A}$): Leveraging the expertise of a LLM, this agent synthesizes the outputs from previous tools along with system environment context to provide 
     a comprehensive fault analysis and recommended solutions.
\end{enumerate}

\subsection{Multi-modality Alignment Tool ($\mathscr{T}_1$)}

Diffusion models are a class of generative models based on Markov processes that have demonstrated exceptional performance in the generation of data in various domains, including images\cite{huang2023collaborativediffusionmultimodalface}, voice\cite{popov2021diffusion}, and time series\cite{yuan2024diffusiontsinterpretablediffusiongeneral}. The core idea is to gradually add noise to the data through a forward denoising (forward diffusion) process, which converts the original data distribution to an approximate Gaussian distribution. In the reverse denoising process, the model learns how to gradually remove the noise to generate the desired data. In this tool, we use log patterns and temporal features as control conditions for two diffusion branches. During the denoising process, the two branches collaborate to guide the generation of time-series data with multimodal features from the noise-perturbed original data. 
Specifically, during the forward diffusion process, the original data $x_0 \sim q(x_0) \in \mathbb{R}^{E\times m\times l} $ consists of $E$ entities, $m$ channels, and a time series of length $l$. This process can be expressed as:
\begin{equation}
\label{eq:diffusion_forward}
    q\left(x_1, \cdots, x_T \mid x_0\right):=\prod_{t=1}^T q\left(x_t \mid x_{t-1}\right).
\end{equation}
If $\alpha_t\in (0,1)$ is a predefined noise scaling factors that determines how much information from the previous step is retained, then each step of the Eq.\ref{eq:diffusion_forward} is:
\begin{equation}
q\left(\mathbf{x}_t \mid \mathbf{x}_{t-1}\right)=\mathcal{N}\left(\mathbf{x}_t ; \sqrt{\alpha_t} \mathbf{x}_{t-1}, \left(1-\alpha_t\right) \mathbf{I}\right).
\end{equation}
After $T$ steps of noise addition, $x_T$ converges to random noise $x_T\sim \mathcal{N}(0,I)$, which approximates an Gaussian distribution. 

Subsequently, we perform stepwise denoising on the noisy data using a dual-branch conditional mechanism. The parameterized conditional transition distribution can be expressed as:
    \begin{align}
    p_\theta\left(\mathbf{x}_{t-1} \mid \mathbf{x}_t,\mathcal{C}\right)&=\mathcal{N}\left(\mathbf{x}_{t-1} ; \boldsymbol{\mu}_\theta\left(\mathbf{x}_t, t, \mathcal{C}\right), \mathbf{\Sigma}_\theta\left(\mathbf{x}_t, t, \mathcal{C}\right)\right), \nonumber \\
     p_\theta\left(\mathbf{x}_{0: T},\mathcal{C}\right)&=p\left(\mathbf{x}_T\right) \prod_{t=1}^T p_\theta\left(\mathbf{x}_{t-1} \mid \mathbf{x}_t ,\mathcal{C}\right),
    \end{align}
where $\mathcal{C}=\{c_{log},c_{time}\}$ represents the set of control conditions, with $c_{log}$ and $c_{time}$  corresponding to the log branch and the time-series branch, respectively. These conditions guide the denoising process, ensuring that the generated data have multi-modal features by utilizing both log-based patterns and temporal features.

For the generation of the log-conditioned variable $c_{log}$, since unstructured log representations are difficult to extract high-quality features, inspired by previous work \cite{han2024holistic} \cite{lee2023eadro}, we extract event patterns $\mathcal{E}$ and keywords $\mathcal{W}$ from the raw logs using Drain\cite{he2017drain}, and by calculating the TF-IDF scores, we select the highest scoring pattern-keyword pairs as the conditional variable $c_{log}$.
\begin{equation}
c_{log}=\left\{\arg \max _{e \in \mathcal{E}} \operatorname{TF-IDF}(e)\right\} \cup\left\{\arg \max _{w \in \mathcal{W}} \operatorname{TF-IDF}(w)\right\}.
\end{equation}

For the generation of the time-conditioned variable $c_{time}$, we extract key performance indicators (KPIs) such as latency and duration from trace files. These KPIs $x_{kpi}$ are then embedded into a latent space using a Multi-Layer Perceptron (MLP) network, enabling the model to effectively incorporate temporal characteristics into the denoising process.
\begin{equation}
c_{time}=\operatorname{Embedding}\left(x_{kpi}\right) \in \mathbb{R}^{h_{{time}}}.
\end{equation}

During training, we use a dual-branch network to learn the reverse diffusion process, gradually denoising the noisy data. In this process, the noise term estimated at each step is collaboratively generated by the two branches.
\begin{equation}
    \begin{aligned}
\mathbf{x}_{t-1}=\frac{1}{\sqrt{\alpha_t}}\left(\mathbf{x}_t-\frac{1-\alpha_t}{\sqrt{1-\bar{\alpha}_t}} \epsilon_\theta\left(\mathbf{x}_t, t,\mathcal{C} \right )\right)+\sigma_t \mathbf{z} \\
 \mathbf{z} \sim \mathcal{N}(0, \mathbf{I}),
    \end{aligned}
\end{equation}
in which, $\bar{\alpha}_t=\prod_{s=1}^t \alpha_s$ represents the cumulative product of the noise scaling factors $\alpha_s$ and $\epsilon_\theta$ represents the noise term predicted by the model, which is jointly generated based on the computations from the log branch $\epsilon_{log}$ and time-series branch $\epsilon_{time}$. A hyperparameter $\mu$ is used as a mixing coefficient to balance the contributions of the two branches. 
\begin{equation}
\epsilon_\theta\left(\mathbf{x}_t, t, \mathcal{C}\right)=\mu \epsilon_{log}\left(\mathbf{x}_t, t, \mathbf{c}_{log}\right)+\left(1-\mu\right) \epsilon_{time}\left(\mathbf{x}_t, t, c_{time}\right).
\end{equation}

The original denoising diffusion model employs U-Net as the denoising backbone \cite{ho2020denoising}. However, considering the characteristics of time-series data, we adopt a patch-based Transformer network \cite{nie2023timeseriesworth64} as the backbone for each branch to better capture the representations of time-series subsequences. 
The training objective of the denoising process is to minimize the KL-divergence between the distributions $p$ and $q$. According to \cite{ho2020denoising}, the objective derivation is formulated as follows:
\begin{equation}
\mathcal{L}_{diff}=\mathbb{E}_{\mathbf{x}_0, \boldsymbol{\epsilon}}\left[\left\|\boldsymbol{\epsilon}-\boldsymbol{\epsilon}_\theta(x_t,t,\mathcal{C})\right\|_2^2\right].
\end{equation}
Through this denoising process, we obtain time series data $x^\prime_{ent} \in \mathbb{R}^{L\times d_{ent}}$ that integrates multimodal features for subsequent analysis, where $ent$ denotes the corresponding resource entity, $L$ denotes the length of the time window, and $d_{ent}$ denotes the dimension of the feature associated with each entity. 
This generative alignment avoids the semantic misalignment inherent in feature concatenation, yielding a more coherent and temporally consistent multimodal representation for downstream analysis ($\mathscr{T}_2$ and $\mathscr{T}_3$).

\subsection{Root Cause Localization Tool ($\mathscr{T}_2$)}
Through the dual-branch diffusion model of Tool $\mathscr{T}_1$, we obtain time-series data that integrates multimodal features. However, since this process operates at the level of system resource entities, the generated time-series $x^\prime_{ent}$ maintains the same dimensionality as the original resource entities.
There exists heterogeneity in the observable data from different resource entities in cloud-native systems, leading to inconsistencies between them, which hinders the unified learning of the model.
To address this issue, we first perform data embedding on the features of different entities, mapping them into a unified dimensional space for consistency:
\begin{equation}
H^{ent}=\operatorname{DataEmbedding}\left(x^\prime_{ent}\right) \in \mathbb{R}^{L \times d_{model}},
\end{equation}
where $d_{model}$ represents the unified embedding dimension. After mapping all entity features into the latent space, we concatenate the output results to obtain the embedded vector $H^{emb}\in \mathbb{R}^{E \times L \times d_{model}}$. 

After obtaining the unified embedding vector $H^{emb}$, we consider that root cause localization tasks typically require the detection and identification of short-term anomalies or disturbance signals within the system entity. These transient features may be obscured by smoothed trends in the time domain. To address this issue, we apply the Fast Fourier Transform (FFT) to convert the data from the time domain to the frequency domain. Formally, the transformation can be expressed as:
\begin{equation}
\mathcal{F}\left(H^{emb}\right)=\operatorname{FFT}\left(H^{emb}\right)=\sum_{i=0}^{L-1} H^{emb}_i e^{-j 2 \pi k i / L}, \quad 
\end{equation}
 where $e^{-j 2 \pi k i / N}$ is the complex exponential fourier basis function, $j$ is the imaginary unit, $H^{emb}_i$ represents the time-series value corresponding to the time step $i$, and $k=0,1, \ldots, N-1$  is the frequency index.

Considering the real-time demands of root cause localization, the model needs to focus on short-term fluctuations in the observed data. To achieve this, we apply a high-pass filter to the frequency-domain data, retaining the first $k$ high-frequency components which helps to amplify the anomalous signals within the data. Formally, this can be expressed as:
\begin{equation}
\operatorname{Indices} = topk\left(\left|\mathcal{F}\left(H^{emb}\right)\right|, k=f r e q \_t o p k\right),
\end{equation}
\begin{equation}
    Mask= \begin{cases}1, & \text { if } k \in \text { Indices } \\ 0, & \text { otherwise }\end{cases},k=0,1, \ldots, K-1,
\end{equation}
\begin{equation}
    \mathcal{F}_{\text {filtered}} =\mathcal{F}\left(H^{emb}\right) \odot Mask, \mathcal{F}_{\text {filtered}}\in \mathbb{R}^{E \times K \times d_{model}},
\end{equation}

Since the interactions between microservices in the cloud-native systems can be described by a dependency graph, in order to fully consider the relationships between entities in the system and capture potential fault propagation pathways, we use the self-attention mechanism to construct a causal dependency graph in the frequency domain. The process can be defined as:
\begin{equation}
    A = Softmax\left(Attention(\mathcal{F}_{\text {filtered}}W^Q,\mathcal{F}_\text {filtered}W^K)\right)
\end{equation}
where $W^Q$,$W^K$ are the query and key matrices respectively. Subsequently, to reduce the influence of irrelevant entities and remove redundant relationships, we introduce topological constraints into the graph generation process. These constraints can be defined as:
\begin{equation}
\mathbf{A}_{\text {maske}}=\mathbf{A} \odot\left(\mathbf{I}-\mathbf{I}_E\right) \odot \operatorname{TopkMask}(\cdot, k=topk)
\end{equation}
in which $\mathbf{I}_E$ is the entity diagonal matrix for avoiding self-loops and $\operatorname{TopkMask}$ preserves the top $k$ most significant dependencies for each entity. Then, we apply graph neural network to perform root cause inference and localization. Specifically, we select the Graph Attention Network (GAT) \cite{brody2021attentive} as the backbone for aggregating neighborhood feature information among system entities, in order to simulate the fault propagation process as follows:
\begin{equation}
    \label{eq:graph1}
    \mathcal{H}^{s+1}_{i}=Relu\left( \alpha_{i i} W \mathcal{H}^s_{j} + \sum_{j \in N(i)} \alpha_{i j} W \mathcal{H}^s_{j} \right),
\end{equation}
where $N(i)=\{j|A_{mask}[i,j]=1\}$, $\mathcal{H}^s_i$ is the representation of entity $i$ in the $s$-th layer and $\mathcal{H}^0_i=\mathcal{F}_{\text {filtered}}$.
$\alpha_{i,j}$ is the attention coefficient between system entity $i$ and its related entity $j$, which can be calculated by the following formula:
\begin{equation}
    \xi(i,j)=LeakyRelu \left( a^\top \left( W^{s}\cdot h^{s}_{i} \oplus W^{s}\cdot h^{s}_{j} \right) \right),
\end{equation}
\begin{equation}
    \label{eq:graph2}
    \alpha_{i j}=\frac{\operatorname{exp} ( \xi(i, j))} {\sum_{p \in N_i \cup\{i \}} \operatorname{exp} ( \xi(i, p))},
\end{equation}
where $\oplus$  denotes the concatenation operation, $a^T$ represents the vector of attention parameters and $W^{s}$ is the weight matrix for the $s$-th layer. Finally, we use a max-pooling layer to aggregate the importance of instances, thereby obtaining the root cause probability for the system entities:
\begin{equation}
\hat{y}=\text { MaxPool}\left(\mathbf{H}^{S}\right) \in \mathbb{R}^{E},
\end{equation}
Considering that multiple types of entities may experience faults simultaneously within the same time period, meaning that the root cause entity may not be unique, we use binary cross-entropy (BCE) as the loss function to optimize the model. Specifically, for each label $y_i$ and the model predicted value $\hat{y_i}$ (where $i\in \{0,1,...,E-1\}$, and $E$ represent the total number of system entities) , the loss function is defined as:
\begin{equation}
    \mathcal{L}_{BCE} = - \frac{1}{E} \sum_{i=1}^{E} \left[ y_i \log(\hat{y_i}) + (1 - y_i) \log(1 - \hat{y_i}) \right],
\end{equation}

\subsection{Fault Types Classification Tool ($\mathscr{T}_3$)}
After the root cause entity localization is performed by Tool $\mathscr{T}_2$, the fault diagnosis process typically requires further identification of the specific fault type of the root cause entity (e.g., hardware overload, resource leakage, or configuration errors) to guide precise remediation strategies. Although root cause localization narrows down the scope of fault analysis (i.e., determining ``where" the anomaly occurs), in practical operational scenarios, different fault types often correspond to differentiated handling strategies (e.g., memory leaks require service restarts, while CPU contention necessitates resource scaling). Therefore, this tool uses the time series data $x'_{ent}$ generated by tool $\mathscr{T}_1$ (for simplicity, denoted as $z$) to train a fault classification model.

To model the latent temporal dependencies in the input time-series data $z$, we use a Transformer model \cite{vaswani2017attention} as the encoder network. The output of its single-head self-attention mechanism is given by:
\begin{equation}
    \text{Attention} = 
    \text{softmax}\left( \frac{zW^{\hat{Q}} \cdot z{(W^{\hat{K}}})^\top}{\sqrt{d_k}} \right) zW^{\hat{V}},
\end{equation}

where $W^{\hat{Q}}$,$W^{\hat{K}}$,$W^{\hat{V}}$ are linear projection matrices. Subsequently, we concatenate the results of the multi-head attention to obtain the latent vector representation:
\begin{equation}
\label{}
    V = 
    \text{MultiHead}\left(zW^{\hat{Q}},zW^{\hat{K}},zW^{\hat{V}}\right),
\end{equation}

After capturing the internal temporal dependencies of the time-series data, we employ the GAT model to capture the spatial dependencies between the time steps. According to formulas Eq.\ref{eq:graph1}-Eq.\ref{eq:graph2}, we perform feature propagation on the latent vector representation. The representation of each entity is obtained by aggregating the weighted features of its neighboring entities:
\begin{equation}
    Z' = \text{ELU}\left(\text{GATConv}(X, A)\right),
\end{equation}
where ELU is the activation function and $A\in \mathbb{R}^{E\times E}$ is the dependency between entities. Subsequently, we use an MLP network to output the fault type for the corresponding entity. Considering the issues of multi-class overlap and class imbalance in the fault classification task, we adopt a weighted binary cross-entropy (wBCE) loss function with the introduction of class weight coefficients $\lambda$ as the model’s training objective:
\begin{equation}
    \mathcal{L}_{wBCE} = - \frac{1}{E} \sum_{i=1}^{E} \left[ \lambda_w \cdot y_i \log(\hat{y_i}) + (1 - y_i) \log(1 - \hat{y_i}) \right],
\end{equation}
in which $\lambda_w$ represents the ratio of negative to positive samples within each fault type class, which is used to enhance the loss contribution of the samples. Additionally, to prevent overfitting, we include an L2 weight decay term in the loss function. The final loss function can thus be formulated as:
\begin{equation}
    \mathcal{L} = \mathcal{L}_{wBCE} + \beta \|\mathbf{W}\|^2_2.
\end{equation}
where $\beta$ is a predefined hyperparameter and $\mathbf{W}$ is the model weights.

\begin{figure}
    \centering

\definecolor{darkred}{rgb}{0.55, 0, 0}
\begin{tcolorbox}[title=RCA Expert Prompt, colback=gray!10, colframe=gray!70!black]
You are an RCA expert responsible for analyzing and diagnosing failures of complex systems based on cloud-native, microservices architectures. You have access to detailed failure reports, system performance data, and historical failure patterns. You provide detailed analysis based on the following structured inputs, which include root cause information, failure classification, system architecture context, historical failure data, and other system state details. You are expected to apply advanced reasoning methods, drawing on knowledge of system architecture, historical failure modes, and industry best practices, to provide comprehensive, practical solutions to improve system reliability and minimize downtime.

[Failure root cause localization]

\quad - Root Entities: \textcolor{darkred}{\textless{}localization results from $\mathscr{T}_2$\textgreater{}}

\quad - Related Entities: \textcolor{darkred}{\textless{}the related calling entities\textgreater{}}

[Fault Classification]

\quad - Fault Type: \textcolor{darkred}{\textless{}classification results from $\mathscr{T}_3$\textgreater{}}

\quad - Symptoms: \textcolor{darkred}{\textless{}detailed Symptom Description\textgreater{}}

[System architecture background]

\quad - Architecture: \textcolor{darkred}{\textless{}microservice functionality\textgreater{}}

\quad - Context: \textcolor{darkred}{\textless{}related logs and trace information\textgreater{}}

[Task Objective] 

\quad \quad Based on the information above, please analyze the root cause, impact scope, and proposed solutions for the fault, and provide the following:

\begin{enumerate}
\item Analysis of the root cause of the fault.
\item Analysis of the type and level of failure.
\item Repair solutions and preventive measures.
\end{enumerate}

\end{tcolorbox}
\caption{Prompt templates for this RCA expert agent.}
\label{fig:LLM-prompt}
\end{figure}

\subsection{RCA Expert Agent ($\mathscr{A}$)}
After the three tools $\mathscr{T}_1$-$\mathscr{T}_3$ have completed tasks such as multi-modality alignment, root cause localization, and fault types classification, the framework has achieved a relatively deep understanding of the fault. However, relying solely on these outputs makes it difficult to provide comprehensive and accurate failure analysis and remediation plans. Therefore, to further improve fault response and system recovery efficiency, we introduce the RCA expert agent $\mathscr{A}$. Using large language models, this agent synthesizes the predictions from the previous tools and incorporates the current context of cloud-native microservice architecture to offer a more comprehensive and accurate fault diagnosis and resolution strategy for site reliability engineers.

Specifically, we use the GPT-4 \cite{achiam2023gpt} as the expert model, which gathers the outputs from the previous tools as background knowledge, including:
\begin{itemize}
    \item \textbf{Root Cause Localization Results} (from $\mathscr{T}_2$): Identifying the root cause location of the fault and the associated system components.
    \item \textbf{Fault Types Classification Results} (from $\mathscr{T}_3$): Providing the specific type of fault, such as network failure, hardware failure, configuration errors, etc.
    \item \textbf{System Context Information}:  Includes the microservice architecture (e.g., service roles, dependencies, and call chains), deployment topology, and resource usage patterns. This context, often derived from system documentation, guides the agent in understanding operational relationships and diagnosing faults.
\end{itemize}

In practice, we prepare the prompt in the format shown in Fig.\ref{fig:LLM-prompt} as input to the model, and the expected output is a detailed fault analysis report along with corresponding solution recommendations. Such outputs assist the engineers in quickly understanding the root cause, impact scope, and taking reasonable remediation actions, thus enhancing system reliability and response efficiency.

\begin{table*}[htb!]
\centering
\caption{Experimental results of different approaches on root cause localization and fault classification. Best results are bolded, and second-best results are underlined}
\label{table:result}
\begin{tabular}{llcccccccccc}
\toprule
\textbf{Data} & \textbf{Approach} & \multicolumn{3}{c}{\textbf{Root Cause Localization}}  & \multicolumn{6}{c}{\textbf{Fault Types Classification}} \\
\cmidrule(lr){3-5}  \cmidrule(lr){6-11}
& & Acc@1 & Acc@3 & Acc@5 & MiPr & MaPr &MiRe &MaRe & MiF1 & MaF1 \\
\midrule
\multirow{10}{*}{$A_s$} 
& HolisticRCA & \underline{65.62\%} & \underline{76.33\%} & \underline{81.69\%} & \underline{0.6355} & \textbf{0.7130} & \textbf{0.7728} & \textbf{0.8037} & \textbf{0.6974} & \textbf{0.7507} \\
& TimesNet & 14.81\% & 32.10\% & 61.72\% & - & -  & - & - & - & - \\
& Eadro & 13.58\% & 37.03\% & 45.68\% & 0.1887 & 0.0770 & 0.1235 & 0.1046 & 0.1493 & 0.0758 \\
& SegRNN & 14.81\% & 40.75\% & 60.49\%  & - & - & - & - & - & - \\
& DejaVu & 62.50\% & 65.62\% & 78.13\%  & - & - & - & - & - & - \\
& LightGBM & - & - & - & 0.1967 & 0.1536 & 0.2000  & 0.2282 & 0.1983 & 0.1707 \\
& Transformer & 16.22\% & 37.84\% & 56.76\% & 0.0114 & 0.0123 & 0.1892 & 0.1932 & 0.0214 & 0.0230 \\
& PatchTST & 14.81\% & 34.57\% & 53.09\% & 0.2000 & 0.0741 & 0.0247 & 0.0185 & 0.0440 & 0.0296 \\
& AutoFormer & 11.11\% & 29.63\% & 37.04\%  & - & - & - & - & - & - \\
& TAMO & \textbf{71.87\%} & \textbf{82.14\%} & \textbf{88.83}\%  & \textbf{0.7164} & \underline{0.6671} & \underline{0.6486} & \underline{0.6149} & \underline{0.6809} & \underline{0.6299} \\
\midrule
\multirow{10}{*}{$A_p$} 
& HolisticRCA & \textbf{65.62\%} & \textbf{80.80\%} & \underline{86.60\%} & 0.5913 & \textbf{0.6682} & \textbf{0.7534} & \textbf{0.7598} & 0.6626 & \textbf{0.7008} \\
& TimesNet & 10.71\% & 22.14\% & 32.14\%  & - & - & - & - & - & - \\
& Eadro & 9.38\% & 13.13\% & 15.63\% & 0.3901 & 0.3344 & 0.4665 & 0.4792 & 0.4249 & 0.3761 \\
& SegRNN & 18.75\% & 56.25\% & 84.38\%  & - & - & - & - & - & - \\
& DejaVu & 18.75\% & 21.88\% & 28.13\%  & - & - & - & - & - & - \\
& LightGBM & - & - & - & \underline{0.6849} & 0.6450 & 0.6792 & \underline{0.6808} & \underline{0.6820} & \underline{0.6583}  \\
& Transformer & 10.71\% & 19.29\% & 24.29\%  & 0.0106 &0.0106 & 0.4254 & 0.4439 & 0.0207 & 0.0208 \\
& PatchTST & 10.00\% & 18.13\% & 23.75\%  & 0.6357 &0.5359 & 0.4069 & 0.4107 & 0.4962 & 0.4543 \\
& AutoFormer & 13.13\% & 25.00\% & 31.25\%  & 0.2600 &0.1602 & 0.1613 & 0.1660 & 0.1991 & 0.1466 \\
& TAMO & \underline{64.28\%} & \underline{80.36\%} & \textbf{87.50\%}  & \textbf{0.7182} & \underline{0.6597} & \underline{0.6840} & 0.6642 & \textbf{0.7007} & 0.6445 \\
\midrule
\multirow{10}{*}{$A_n$} 
& HolisticRCA & 73.66\% & 75.89\% & 79.01\% & \underline{0.6219} & \underline{0.6437} & 0.7612 & 0.7322 & \underline{0.6846} & \underline{0.6717} \\
& TimesNet & 21.15\% & 48.07\% & 78.85\%  & - & - & - & - & - & - \\
& Eadro & 17.18\% & 28.13\% & 42.19\% & 0.5426 & 0.6003 & 0.7969 & 0.8145 & 0.6456 & 0.6575 \\
& SegRNN & 7.50\% & 20.63\% & 23.75\%  & - & - & - & - & - & - \\
& DejaVu & \underline{81.25\%} & \underline{90.63\%} & \textbf{100.00\%}  & - & - & - & - & - & - \\
& LightGBM & - & - & - & 0.5208 & 0.5491 & \underline{0.8333} & \underline{0.8178} & 0.6410 & 0.6406  \\
& Transformer & 13.46\% & 48.08\% & 78.85\%  & 0.1415 & 0.1423 & 0.5577 & 0.5207 & 0.2257 & 0.2212 \\
& PatchTST & 17.19\% & 42.19\% & 79.69\%  & 0.3521 & 0.3597 & 0.3906 & 0.3609 & 0.3704 & 0.3476 \\
& AutoFormer & 21.88\% & 59.38\% & 92.19\%  & 0.5075 & 0.5286 & 0.5313 & 0.5195 & 0.5191 & 0.5181 \\
& TAMO & \textbf{84.37}\% &\textbf{91.96}\% & \underline{97.77\%}  & \textbf{0.8718} & \textbf{0.8869} & \textbf{0.8947} & \textbf{0.8681} & \textbf{0.8831} & \textbf{0.8706} \\
\midrule
\multirow{10}{*}{$B$} 
& HolisticRCA & \underline{61.11\%} & \underline{75.00\%} & 77.78\% & 0.5000 & 0.5936 & \textbf{0.8056} & \textbf{0.8056} & 0.6170 & \textbf{0.6636} \\
& TimesNet & 15.62\% & 25.00\% & 40.63\%  & 0.4893 & 0.5789 & 0.7187 & 0.7398 & 0.5823 & 0.6394 \\
& Eadro & 40.62\% & 56.25\% & 71.87\% & 0.4167 & 0.5211 & 0.6250 & 0.6516 & 0.5000 & 0.5694 \\
& SegRNN & 18.75\% & 43.75\% & 62.50\%  & 0.1496 & 0.2359 & 0.5937 & 0.6287 & 0.2389 & 0.3296 \\
& DejaVu & 15.62\% & 34.38\% & 50.00\%  & - & - & - & - & - & - \\
& LightGBM & - & - & - & \underline{0.7832} & 0.6434 & 0.4186  & 0.3860 & 0.4687 & 0.4122  \\
& Transformer & 46.88\% & 68.75\% & \underline{84.38\%}  & 0.4464 & 0.5664 & \underline{0.7813} & \underline{0.7955} & 0.5682 & 0.6406 \\
& PatchTST & 12.50\% & 34.38\% & 53.13\%  & 0.7200 & \underline{0.7222} & 0.5625 & 0.5934 & \underline{0.6316} & 0.6417 \\
& AutoFormer & 46.88\% & 62.50\% & 75.00\%  & 0.5185 & 0.6177 & 0.4375 & 0.4705 & 0.4746 & 0.5013 \\
& TAMO & \textbf{72.22\%} & \textbf{80.55\%} & \textbf{86.11\%}  & \textbf{0.8500} & \textbf{0.8750} & 0.5313 & 0.5682 & \textbf{0.6538} & \underline{0.6421} \\
\bottomrule
\end{tabular}
\end{table*}

\section{Evaluation}
\subsection{Research Questions}
This section answers the following research questions:
\begin{itemize}
    \item \textbf{RQ1:} How effective is TAMO in Root Cause Localization?
    \item \textbf{RQ2:} How effective is TAMO in Fault Types Classification?
    \item \textbf{RQ3:} How well does TAMO work when ablation studies are performed on its various components?
    \item \textbf{RQ4:} What are the training and inference efficiency of TAMO, and how do hyperparameters affect its performance?
    \item \textbf{RQ5:} How well does TAMO perform in real-world case studies and fault scenarios?
    
\end{itemize}

\subsection{Experiment Setup}
\subsubsection{Datasets}
We conducted extensive experiments on two public datasets: 
\begin{itemize}
    \item \textbf{Dataset} $A$. $A$ is a large-scale public dataset from the AIOps Challenge\footnote{\url{https://competition.AIOps-challenge.com/home/competition/1496398526429724760}}, which is collected through fault injection into the real-world deployed microservice system HipsterShop\footnote{\url{https://github.com/GoogleCloudPlatform/microservices-demo}}. The dataset primarily consists of three types of data: metrics, logs, and traces. The system uses a dynamic deployment architecture, consisting of 10 services, each with 4 pods, totaling 40 pods that are dynamically deployed across 6 nodes. The dataset includes 15 types of faults in total. Among these, 9 fault types are related to services and pods in the context of Kubernetes (K8s) containers \cite{burns2016borg}, while node faults include 6 types: sudden memory pressure, disk space exhaustion, disk read/write issues, CPU pressure, and slow CPU growth. 
    \item \textbf{Dataset} $B$. $B$ is a public dataset\footnote{\url{https://doi.org/10.5281/zenodo.7615393}} collected from the broadcast-style SocialNetwork system in \cite{lee2023eadro}, which only contains service-level entities. It includes 21 microservices and also consists of three types of data: metrics, logs, and traces. The dataset is generated by injecting faults into the system using Chaosblade, producing data that includes three types of faults: CPU resource exhaustion, network latency, and packet loss.
\end{itemize}

\subsubsection{Baselines}
We compare TAMO with the most popular integrated multimodal methods and derivative methods from the time series domain. Specifically, we select two state-of-the-art integrated multimodal root cause analysis methods, Eadro\cite{lee2023eadro} and HolisticRCA\cite{han2024holistic}, as well as four unimodal time series analysis methods (i.e., TimesNet\cite{wu2022timesnet}, SegRNN\cite{lin2023segrnn}, DejaVu\cite{li2022actionable}, Transformer\cite{vaswani2017attention}). These methods are implemented using their publicly available and reproducible open-source code, and we adjust the data format to match the required input. Furthermore, since our research task is relatively novel and includes both fault localization and classification, we incorporate the same fault classifier used in TAMO for methods that do not natively implement fault classification (such as methods \cite{lee2023eadro}, \cite{vaswani2017attention}, \cite{wu2022timesnet}, etc.) for comparison. Furthermore, we introduce LightGBM\cite{ke2017lightgbm}, a method specifically designed for fault classification tasks, as a benchmark to evaluate TAMO performance.

\begin{table*}[htb!]
\centering
\caption{The experimental results of the ablation study conducted on our proposed TAMO framework.}
\label{table:ablation}
\begin{tabular}{lcccccccccc}
\toprule
& \multirow{2}{*}[-0.5ex]{\textbf{Models}}   & \multicolumn{3}{c}{\textbf{Root Cause Localization}}  & \multicolumn{6}{c}{\textbf{Fault Types Classification}} \\
\cmidrule(lr){3-5}  \cmidrule(lr){6-11}
&  & Acc@1 & Acc@3 & Acc@5 & MiPr & MaPr &MiRe &MaRe & MiF1 & MaF1 \\
\midrule
& $\operatorname{TAMO}_{w/o \; \mathscr{T}_1 }$ & 43.75\% & 56.25\% & 68.75\% & 0.5909 & 0.5556 & 0.4063 & 0.4453 & 0.4815 & 0.4911 \\
& $\operatorname{TAMO}_{w/o \; branch }$ & 59.38\% & 68.75\% & 71.87\% & 0.7200 & 0.7350  & 0.5625 & \underline{0.6010} & \underline{0.6316} & 0.6167 \\
& $\operatorname{TAMO}_{w/o \; FFT}$ & 53.13\% & 68.75\% & \underline{78.13\%} & 0.8000 & 0.7639  & 0.5000 & 0.5404 & 0.6154 & 0.5977\\
\midrule
& $\operatorname{TAMO}$ & \textbf{72.22\%} & \textbf{80.55\%} & \textbf{86.11\%}  & \textbf{0.8500} & \textbf{0.8750} & \textbf{0.5313} & \textbf{0.5682} & \textbf{0.6538} & \textbf{0.6421} \\
\bottomrule
\end{tabular}
\end{table*}

\subsubsection{Evaluation metrics}
TAMO is dedicated to locating root cause instances and identifying fault types. Based on previous research works\cite{han2024holistic}\cite{han2024potential}\cite{lee2023eadro}\cite{zhang2023robust}, we select different evaluation metrics to evaluate the performance of model for these two tasks. Specifically, for the root cause localization task, we use top-k accuracy (Acc@K) as the evaluation metric, which represents the probability that the true root cause entity appears within the top-k results produced by the method. Here, we set $k=\{1,3,5\}$. Given $S$ as the set of faults, $RC$ is the ground truth root cause, and $rc[k]$ are the top-k root causes generated by the model. Therefore, Acc@K can be defined as follows:
\begin{equation}
    Acc@k = \frac{1}{|S|} \sum_{i \in S}^S 
    \begin{cases} 
        1, & \text{if } RC_i \in rc_i[k] \\
        0, & \text{otherwise}
    \end{cases}
\end{equation}
For the fault classification task, we use $Precision(Pre)$, $Recall(Rec)$, and $F1-score(F1)$ to evaluate performance. Given true positives (TP), false positives (FP), and false negatives (FN), the formulas for these metrics are as follows: $Pre=\frac{TP}{TP+FP}$,$Rec=\frac{TP}{TP+FN}$, $F1=\frac{2\cdot Pre \cdot Rec}{Pre+Rec}$. Considering that this task is a multi-label classification problem, we adopt both micro and macro metrics for a comprehensive evaluation. Micro metrics aggregate the performance across all labels, treating all instances equally, which is useful when the class distribution is imbalanced. In contrast, macro metrics compute the metric for each label independently and then average the results, giving equal weight to each label regardless of its frequency. Thus, the final performance is measured using micro-precision ($MiPr$), macro-precision ($MaPr$), micro-recall ($MiRe$), macro-recall ($MaRe$), micro-F1 score ($MiF1$), and macro-F1 score ($MaF1$).

\subsubsection{Implementation}
We implement TAMO using Python torch 2.4.0 and CUDA 12.1. The experiments are conducted on a Linux server equipped with an NVIDIA GeForce RTX 3090 GPU, using Python 3.8. The model is trained with the Adam optimizer \cite{kingma2014adam}, and all experiments use a fixed random seed. For hyperparameters, we set the batch size to 32, $\mu$ to 0.5, $\beta$ to 0.001, $epoch$ to 200, and the initial learning rate to 0.001. 
Considering the requirements of TAMO for system contextual awareness and temporal consistency, data cleaning is performed on datasets A and B before training. This process remove noisy and misaligned multimodal records and realigned log and trace data to ensure temporal consistency and enhance the quality of input observation data for $\mathscr{T}_1$. Additionally, considering the pre-training needs of $\mathscr{T}_1$, an extra dataset of observation data from normal runtime operations is constructed. This dataset is used for $\mathscr{T}_1$ pre-training to enable learning of normal system patterns.

For training, we adopt a sequential and modular strategy to ensure stable learning across the tool ensemble. We first pre-train tool $\mathscr{T}_1$ using normal operational data from Datasets A and B, leveraging the diffusion framework to learn robust representations of multi-modal system observations. Once $\mathscr{T}_1$ is pre-trained, its parameters are frozen and it serves as a fixed feature extractor for subsequent stages. Tool $\mathscr{T}_2$ is then trained using the embeddings generated by $\mathscr{T}_1$ to perform root cause localization. After $\mathscr{T}_2$ is trained, both $\mathscr{T}_1$ and $\mathscr{T}_2$ are kept frozen while training $\mathscr{T}_1$-generated embeddings and the causal graphs from $\mathscr{T}_2$ as input for fault classification. Finally, all components are jointly fine-tuned for 5 epochs with a small learning rate of 0.0001 to refine the overall pipeline. This staged training ensures effective knowledge transfer and prevents instability during end-to-end optimization.

Additionally, due to the heterogeneity of system entities in Dataset $A$, which includes three levels: Pod, Service, and Node. The collected data features vary across these levels, making it challenging to train certain baselines uniformly. For such baselines, we partition Dataset $A$ into subsets composed of different entity types (e.g., $A_p$, $A_s$, and $A_n$) and perform separate training and testing for each subset to evaluate their performance.  
Finally, the symbol '-' is used in the results table \ref{table:result} to indicate cases where a method failed to achieve meaningful classification performance (e.g., accuracy close to 0) on different datasets.  

\begin{figure*}[!t]
\centering
\includegraphics[width=0.98\textwidth]{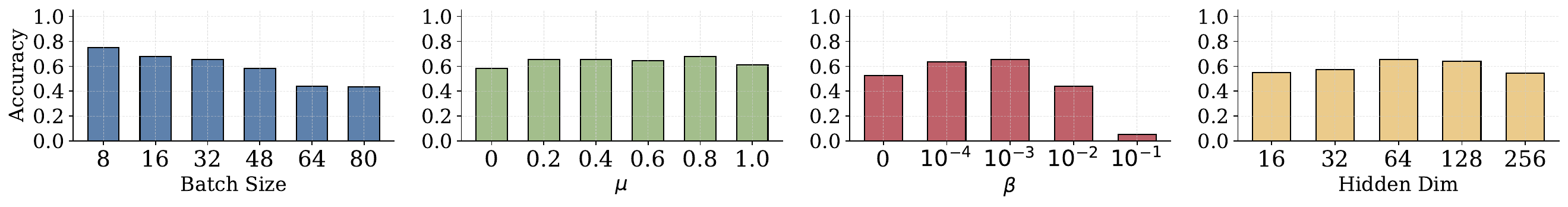}
\caption{Hyperparameter sensitivity analysis across batch size, $\mu$, regularization strength ($\beta$), and hidden dimension.}
\label{fig:hyperparameter}
\end{figure*}

\subsection{\textbf{RQ1:} Effectiveness in Root Cause Localization}
Table \ref{table:result} presents the results for root cause localization. The experimental data shows that our proposed TAMO model achieves the best or second-best performance across all datasets. Specifically, on the $A_s$ dataset, the score of TAMO exceeds the second-best score by at least 6\%, and even in the second-best performance case, the gap with the best result is kept under 3\%, clearly outperforming other baselines. Furthermore, we observe that the Eadro and Transformer models, which perform well on the B dataset, experience a significant drop in performance on heterogeneous datasets such as $A_s$, $A_p$, and $A_n$. This is because these two methods can only handle homogeneous entity features (e.g., in the dataset B, there is only one type of service entity). When confronted with heterogeneous entity features that need to be processed separately, their performance significantly deteriorates due to the lack of inter-entity correlation information. In contrast, both HolisticRCA and our approach model heterogeneous entities in a unified manner, maintaining strong performance even on the dataset A. Compared to HolisticRCA, we map multimodal information into time-series features and further optimize it through spatiotemporal modeling, achieving better performance in the root cause localization task.

\subsection{\textbf{RQ2:} Effectiveness in Fault Types Classification}
In the fault classification task, as shown in Table \ref{table:result}, TAMO demonstrates significant improvments in two key metrics: Micro Precision (MiPr) and Micro F1 (MiF1). 
Particularly on the $A_n$ dataset, TAMO achieves a MiPr of $0.8718$ and MiF1 of $0.8831$, outperforming the second-best model (HolisticRCA) by $24.99\%$ and $19.85\%$, respectively, and ranks first across all evaluation metrics.
With MiPr reaching $0.7164$ and $0.7182$ on $A_s$ and $A_p$, respectively, TAMO outperforms the best baseline by $8.09\%$ and $12.69\%$, significantly reducing the misclassification. 
Furthermore, we observe that transformer-based models achieve relatively strong performance on relatively simpler dataset B.  
However, their performance degrades significantly on the large-scale, complex system dataset A, indicating that metric-only transformer architectures 
struggle to capture the intricate, multi-modal dependencies inherent in large-scale distributed systems.

\begin{figure}[!t]
\centering
\includegraphics[width=0.48\textwidth]{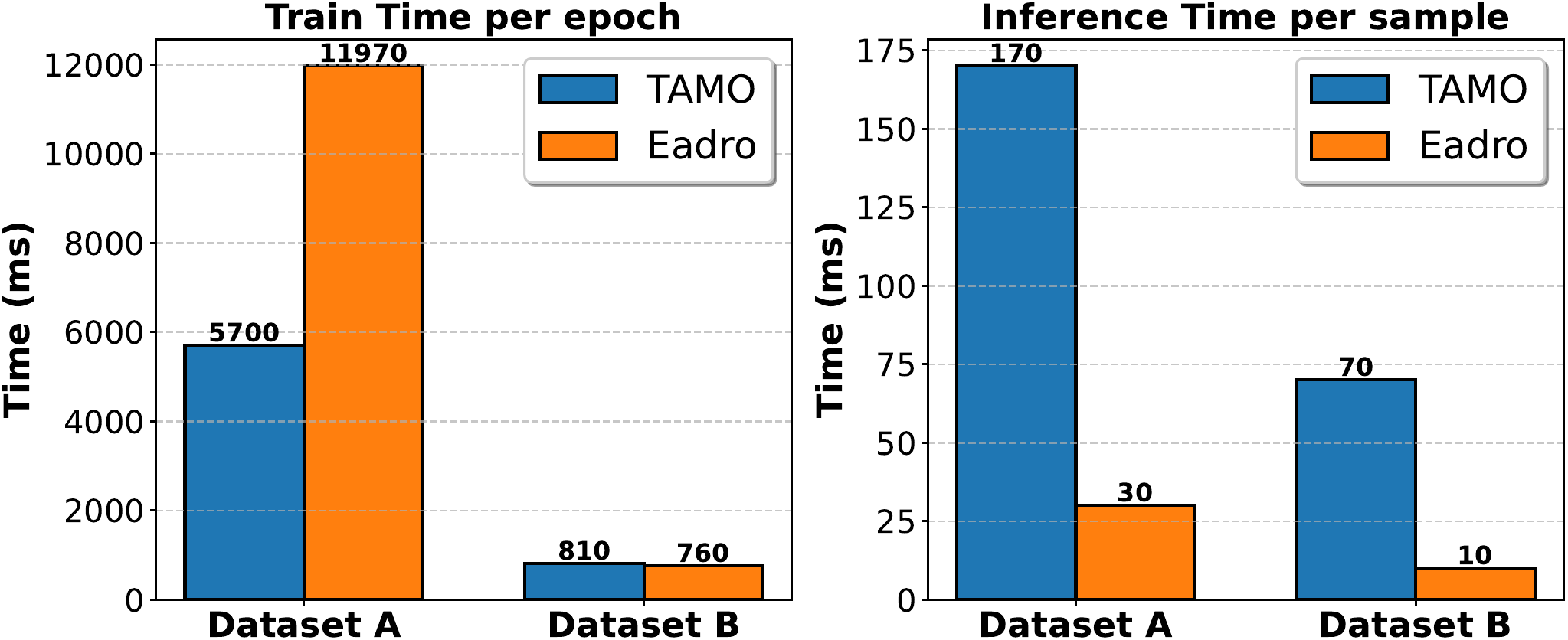}
\caption{Comparision of training time and inference time of TAMO with Eadro. 
}
\label{fig:time_compare}
\end{figure}

\subsection{\textbf{RQ3:} Ablation Study}
To explore the effectiveness of each component in our TAMO framework, we designed three variant models and conducted ablation experiments. The variant models are as follows:
\begin{itemize}
    \item $\operatorname{TAMO}_{w/o \; \mathscr{T}_1 }$: To investigate the effectiveness of the diffusion method used by tool $\mathscr{T}_1$ in fusing multimodal feature, in this variant we remove the tool $\mathscr{T}_1$ and directly use the raw time series data as input to evaluate the variant performance.
    \item $\operatorname{TAMO}_{w/o \; branch }$: To investigate the effectiveness of the dual-branch diffusion model in tool $\mathscr{T}_1$, we remove the branch $\epsilon_{time}$, which is responsible for maintaining temporal features, in this variant. Only the log branch $\epsilon_{log}$ is used in the diffusion reconstruction process. The reconstructed data are then used for fault classification to evaluate variant performance.
    \item $\operatorname{TAMO}_{w/o \; FFT}$: To investigate the effectiveness of the frequency domain-based root cause localization method proposed for tool $\mathscr{T}_2$, we remove the Fourier process in this variant. Instead, the original time-domain data are directly used for subsequent causal graph generation and graph convolution processes. The root cause localization results are then used to evaluate variant performance.
\end{itemize}

Table \ref{table:ablation} presents the results of the ablation experiments, from which it is evident that the removal of any component in TAMO leads to performance degradation. This demonstrates that each of the component within TAMO has an important effect. In particular, the variant of TAMO without tool $\mathscr{T}_1$ shows a significant decrease in performance. This is because the fact that the variant $\operatorname{TAMO}_{w/o \; \mathscr{T}_1 }$ lacks the ability of the diffusion model to fuse multi-modality information, thereby being reduced to learn with only a single modal feature. The absence of key feature information makes accurate root cause analysis challenging. Similarly, removing the time branch $\epsilon_{time}$ lead to noticeable decreases in performance for both tasks, indicating that time series features are crucial for accurate fault localization and classification. Eliminating the FFT process result in a substantial degradation in RE localization metrics, demonstrating that frequency domain analysis is essential for precise root cause localization.

\subsection{\textbf{RQ4:} Performance Study}
To evaluate computational efficiency, we measure the training and inference time of TAMO on both datasets, as shown in Figure~\ref{fig:time_compare}, with Eadro as a representative baseline.
For training, TAMO requires 0.81 seconds per epoch on Dataset B and 5.7 seconds on Dataset A, indicating favorable scalability with minimal computational overhead relative to Eadro.
For inference, TAMO processes each sample in 0.17 seconds, which is slightly slower than Eadro but still meets the real-time requirement for root cause analysis.

In addition, we provide the parameter counts for each component of the TAMO framework in Table~\ref{tab:tool_model_params}. The multi-modality alignment tool $\mathscr{T}_1$ accounts for the majority of parameters (9.32M),  as aligning heterogeneous multimodal data into a unified temporal representation requires a large number of parameters to learn effective alignment. In contrast, the root cause localization tool $\mathscr{T}_2$ and fault types classification tool $\mathscr{T}_3$ are significantly more lightweight (1.38M and 258.32K parameters, respectively).

\begin{table}[h]
\centering
\caption{Parameter counts of the tool models in TAMO.}
\label{tab:tool_model_params}
\begin{tabular}{cc}
\hline
\textbf{Model} & \textbf{Parameters} \\
\hline
  Multi-modality Alignment Tool ($\mathscr{T}_1$) & 9.32M \\
  Root Cause Localization Tool ($\mathscr{T}_2$) & 1.38M \\
  Fault types Classification Tool ($\mathscr{T}_3$) & 258.32K \\
\hline
\textbf{Total parameters}  & \textbf{10.96M} \\
\hline
\end{tabular}
\vspace{0.5em}
\end{table}

Finally, we analyze sensitivity to key hyperparameters  (batch size, $\beta$, $\mu$, hidden dimension). As shown in Fig.~\ref{fig:hyperparameter}, performance is highly sensitive to $\beta$, with minor changes causing sharp accuracy drops. In contrast, smaller batch sizes consistently yield better results, while $\mu$ causes significant performance degradation at its boundary values ($\mu$ = 0 or 1), indicating that it is designed to achieve critical branch balancing.

\begin{figure}[!t]
\centering
\includegraphics[width=0.48\textwidth]{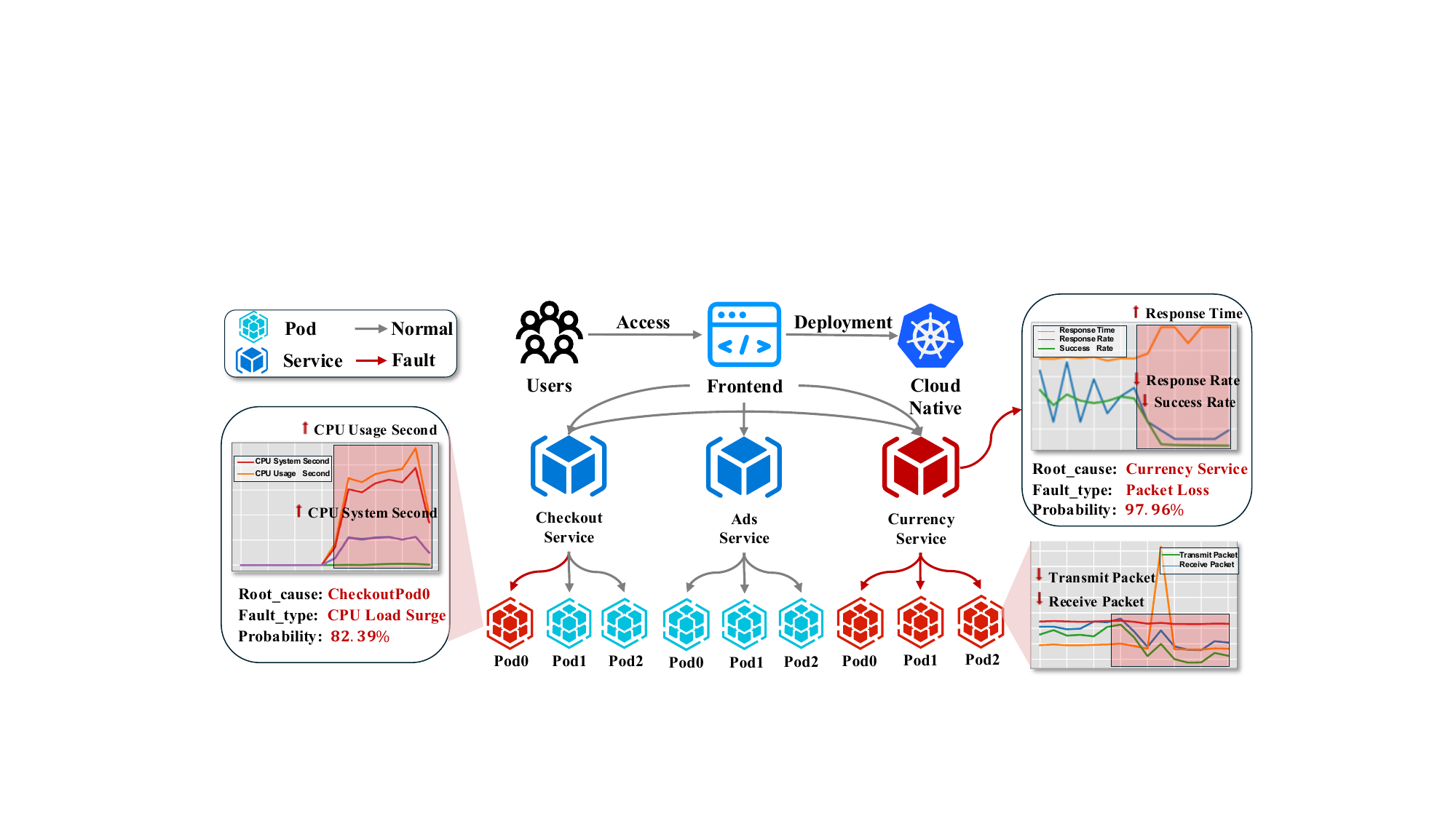}
\caption{In case study experiments, two types of faults were induced in microservices, with the root cause entities identified as the Currency Service and CheckoutPod0.}
\label{fig:case}
\end{figure}

\begin{figure*}[!t] 
\includegraphics[width=0.98\textwidth]{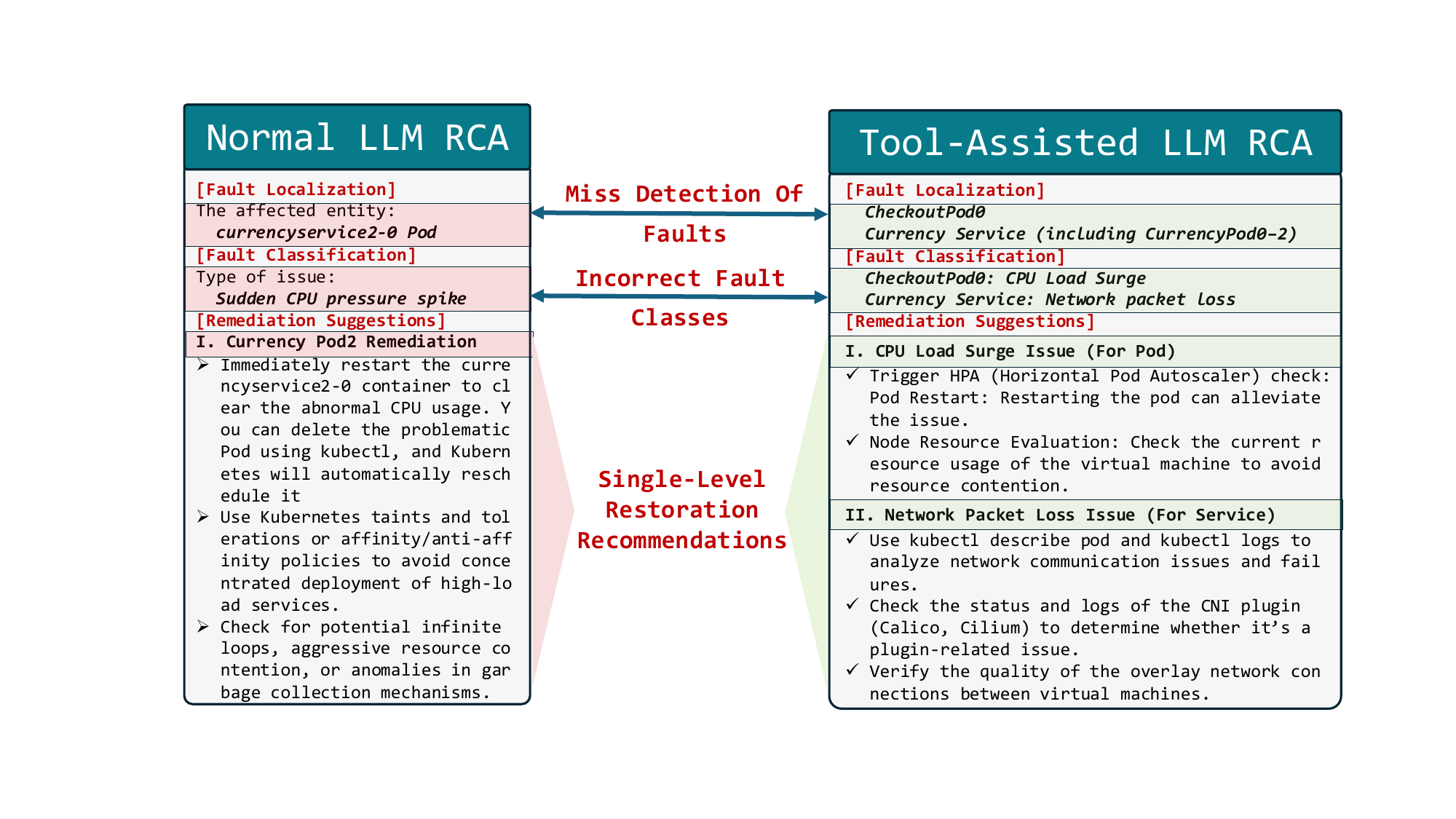}
\caption{Comparison of LLM results between RCA analysis of Agent $\mathscr{A}$ enhanced with TAMO tools and root cause analysis performed directly on raw data}
\label{fig:LLM}
\end{figure*}

\subsection{\textbf{RQ5:} Case Study}
To validate the root cause analysis capabilities of TAMO in cloud-native ecosystems, this study conducts a case investigation using real-world failure data from HipsterShop. As shown in Fig. \ref{fig:case}, the microservices system is deployed using a Kubernetes (K8s) architecture, with service components developed in multiple programming languages. The system entry is managed through the Frontend service, and each service node is configured with three Pod replicas. Monitoring data collection points are set at both the service and Pod levels. Experimental data shows that TAMO demonstrates precise root cause identification for the current failure scenario: the root cause localization confidence for the Currency Service reaches 97.96\% (service level), while the confidence for CheckoutPod0 is 82.39\% (Pod level). Notably, while the overall Checkout service is functioning correctly, only Pod0 experiences issues. In contrast, the Currency service suffers from a global failure at the service level, causing anomalies in Pods 0-2. This comparative case indicates that TAMO effectively integrates multi-entity features and topological causal relationships to achieve precise root cause diagnosis across multiple layers. In the fault classification task, we perform a visual analysis of observation data for the anomalous services/Pods: CheckoutPod0 is accurately classified as a CPU overload fault due to a significant spike in core metrics such as "CPU Usage Second." The Currency service exhibits a correlation between increased service-level response time and decreased request success rate, while its Pod-level Receive Packet count continues to decline. TAMO successfully identifies the anomaly as a network packet loss fault through multi-dimensional time-series pattern analysis. 

Subsequently, we input the model inference results along with system context information into the RCA expert LLM and compare its output with that of an LLM fed directly with raw data, as shown in Fig.\ref{fig:LLM}. This framework leverages tool information and domain expertise to accurately analyze current system faults and provide reasonable remediation suggestions. In contrast, the normal LLM using only raw data struggles to handle the vast amount of observation data due to severe context limitations, resulting in significant omissions and misjudgments in root cause identification and fault classification, as well as corresponding errors in the proposed remediation suggestions.

\section{Conclusion}
In this paper, we introduced a tool-assisted LLM agent framework named TAMO to address the complex challenges of root cause analysis in cloud-native systems. By integrating domain-specific tools with large language models, TAMO efficiently handled high-dimensional, multi-modal cloud-native observation data and accurately located root causes of failures in dynamically changing service dependencies. The approach not only overcame the information loss issues incurred when converting raw observability data into a format suitable for LLM processing, but also addressed the challenge that static pre-trained knowledge struggled to adapt to real-time changes in service dependencies.
The experimental results show that, compared to existing advanced methods and benchmarks, TAMO performs notably better on two standard datasets, highlighting its potential to improve fault diagnosis efficiency and accuracy. 
\bibliography{mybibliography}
\bibliographystyle{plain}


\begin{IEEEbiography}[{\includegraphics[width=1in,height=1.25in,clip,keepaspectratio]{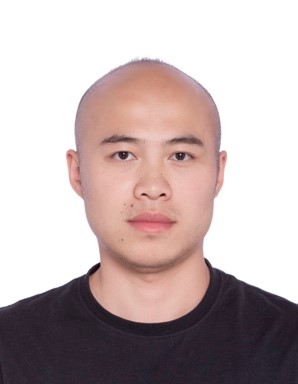}}]{Xiao Zhang} is now an associate professor in the School of Computer Science and Technology, Shandong University. His research interests include data mining, distributed  learning and edge intelligence. He has published more than 60 papers in the prestigious refereed journals and conference proceedings, such as IEEE Transactions on Knowledge and Data Engineering, IEEE Transactions on Mobile Computing, ICML, NeurIPS, SIGKDD,  UBICOMP, and INFOCOM.  
\end{IEEEbiography}

\begin{IEEEbiography}[{\includegraphics[width=1in,height=1.25in,clip,keepaspectratio]{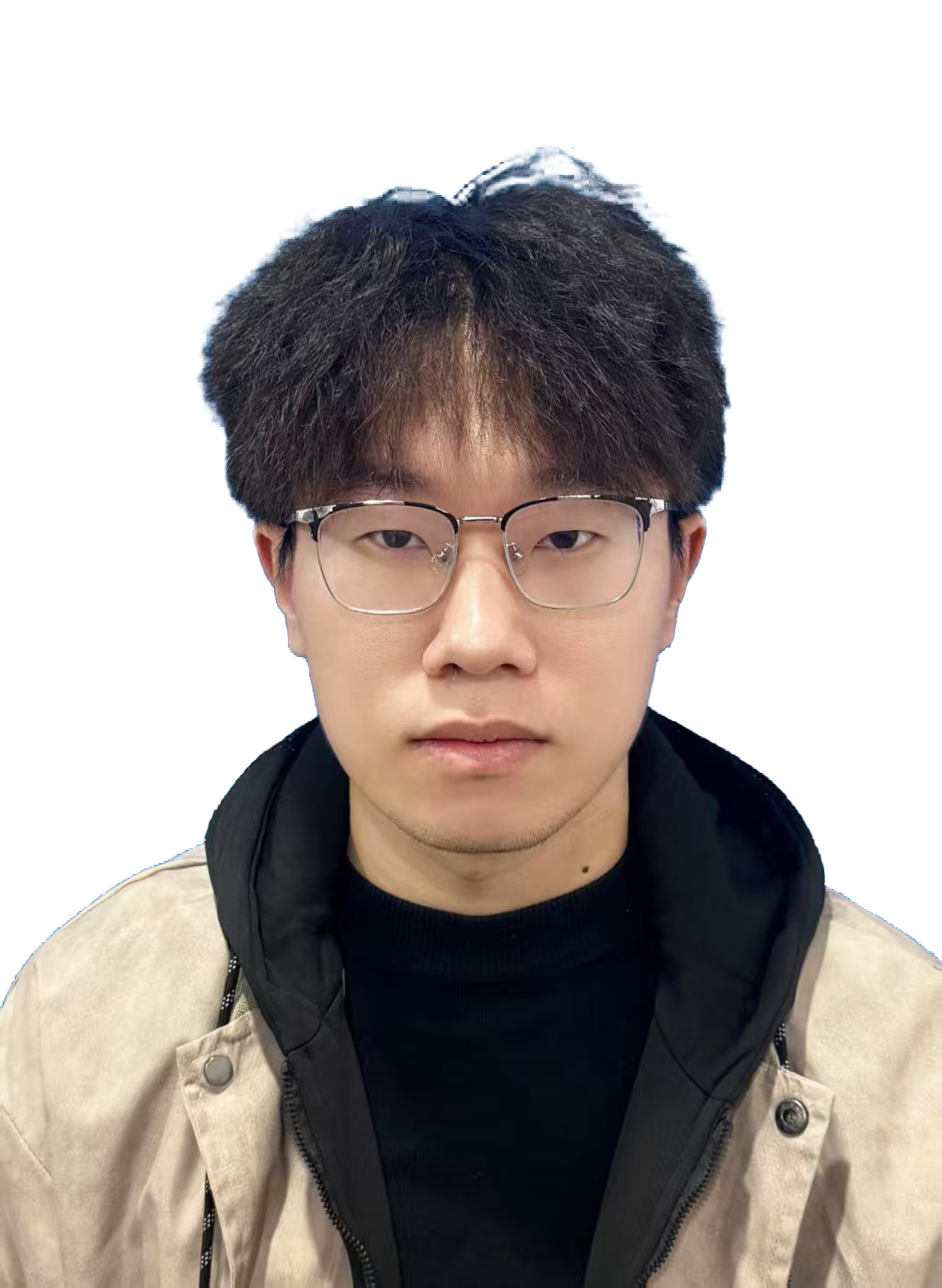}}]{Qi Wang} 
 is a M.D. student in the School of Computer Science and Technology at Shandong University. He received his B.S. degree from Shandong University. His current research interests include time series and predictive maintenance. 
\end{IEEEbiography}

\begin{IEEEbiography}[{\includegraphics[width=1in,height=1.25in,clip,keepaspectratio]{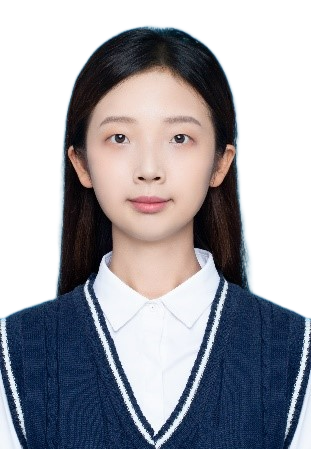}}]{Mingyi Li} is currently a Ph.D. student in the School of Computer Science and Technology, Shandong University. She received her B.S. degree in Shandong University. Her research interests include distributed collaborative Learning and the theoretical optimization of distributed algorithms.
\end{IEEEbiography}

\begin{IEEEbiography}[{\includegraphics[width=1in,height=1.25in,clip,keepaspectratio]{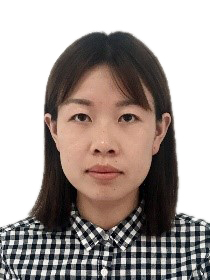}}]{Yuan Yuan} received the BSc degrees from the School of Mathematical Sciences, Shanxi University in 2016, and the Ph.D. degree from the School of  Computer Science and Technology, Shandong University, Qingdao, China, in 2021. She is currently a postdoctoral fellow at the Shandong University-Nanyang Technological University International Joint Research Institute on Artificial Intelligence, Shandong University. Her research interests include distributed computing and distributed machine learning.
\end{IEEEbiography}

\begin{IEEEbiography}[{\includegraphics[width=1in,height=1.25in,clip,keepaspectratio]{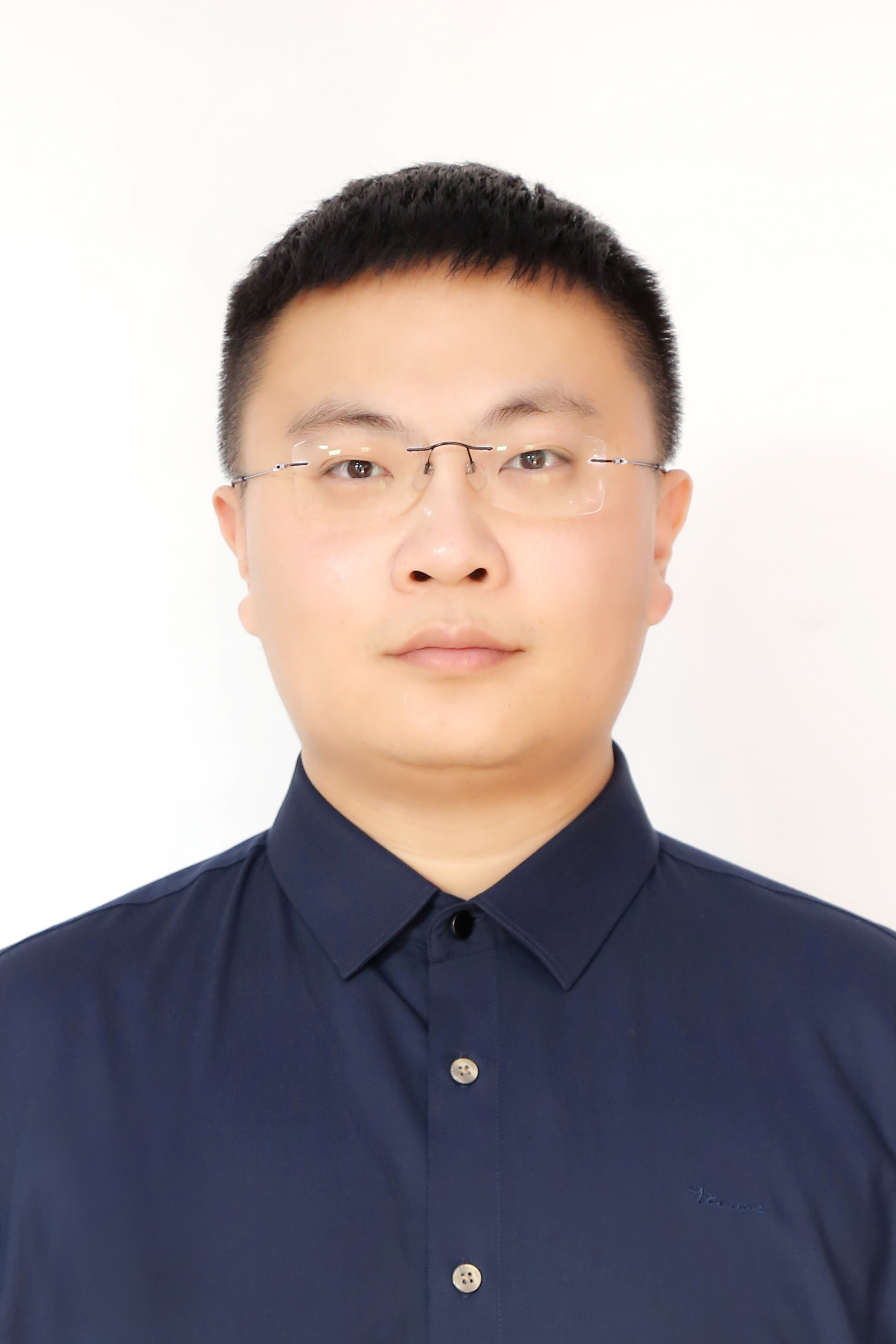}}]{Mengbai Xiao}, Ph.D., is a Professor in the School of Computer Science and Technology at Shandong University, China. He received the Ph.D. degree in Computer Science from George Mason University in 2018, and the M.S. degree in Software Engineering from University of Science and Technology of China in 2011. He was a postdoctoral researcher at the HPCS Lab, the Ohio State University. His research interests include multimedia systems, parallel and distributed systems. He has published papers in prestigious conferences such as USENIX ATC, ACM Multimedia, IEEE ICDE, IEEE ICDCS, IEEE INFOCOM.
\end{IEEEbiography}

\begin{IEEEbiography}[{\includegraphics[width=1in,height=1.25in,clip,keepaspectratio]{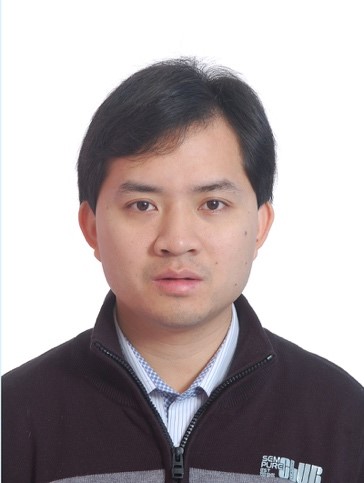}}]{Fuzhen Zhuang} received the Ph.D. degrees in the Institute of Computing Technology, Chinese Academy of Sciences, Beijing, China, in 2011. He is currently a full Professor with the Institute of Artificial Intelligence, Beihang University. He has published more than 150 papers in some prestigious refereed journals and conference proceed-ings such as Nature Comunications, the IEEE Transactions on Knowledge and Data Engineering, the IEEE Transactions on Cybernetics, the IEEE Transactions on Neural Networks and Learning Systems, the ACM Transactions on Knowledge Discovery from Data, the ACM Transactions on Intelligent Systems and Technology, Information Sciences, Neural Networks, SIGKDD, IJCAI, AAAI, TheWebConf, ACL, SIGIR, ICDE, ACM CIKM, ACM WSDM, SIAM SDM, and IEEE ICDM. His research interests include transfer learning, machine learning, data mining, multitask learning, knowledge graph and recommendation systems. He is a Senior Member of CCF. He was the recipient of the Distinguished Dissertation Award of CAAI in 2013.
\end{IEEEbiography}

\begin{IEEEbiography}[{\includegraphics[width=1in,height=1.25in,clip,keepaspectratio]{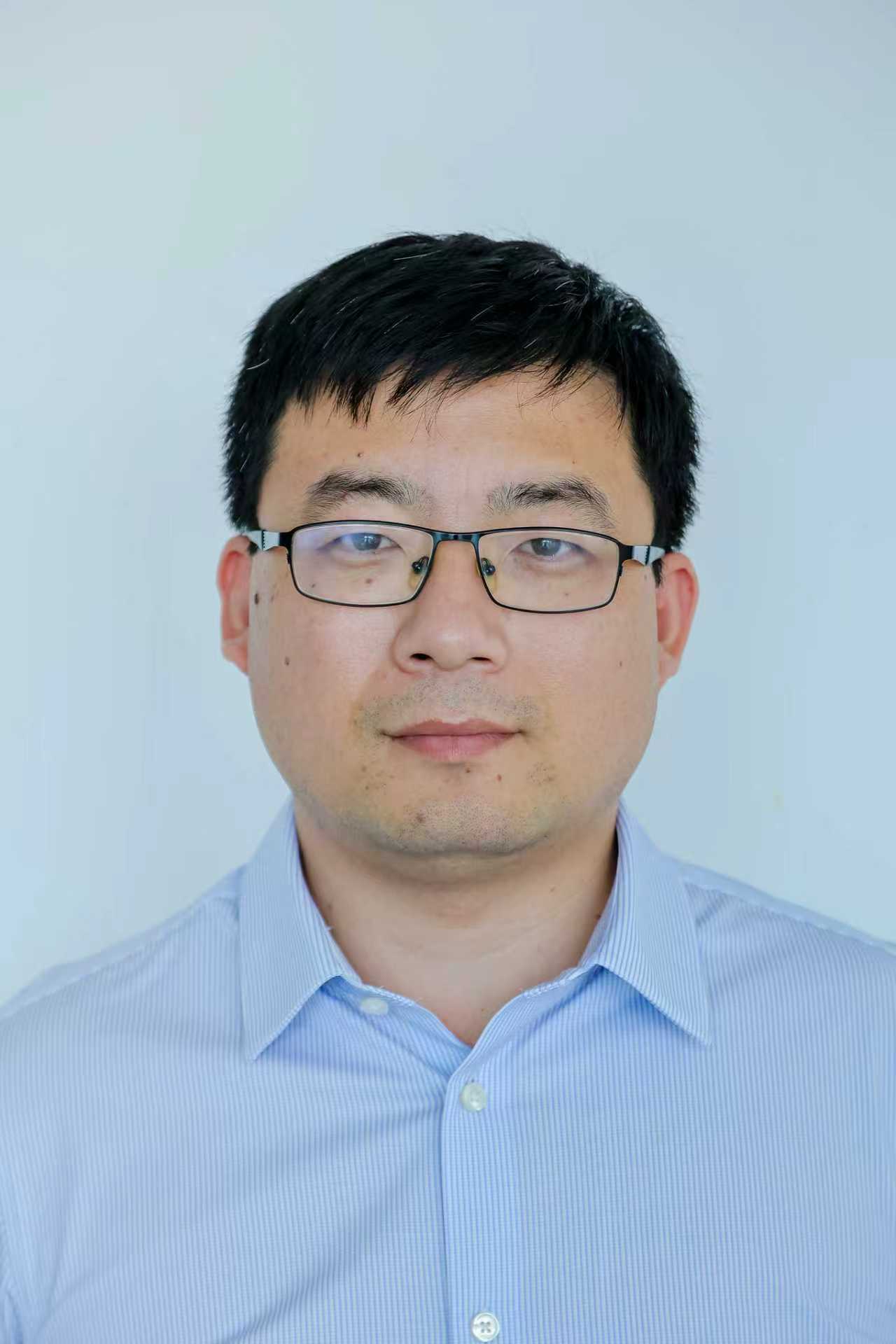}}]{Dongxiao Yu} received the B.S. degree in 2006 from the School of Mathematics, Shandong University and the Ph.D degree in 2014 from the Department of Computer Science, The University of Hong Kong. He became an associate professor in the School of Computer Science and Technology, Huazhong University of Science and Technology, in 2016. He is currently a professor in the School of Computer Science and Technology, Shandong University. His research interests include edge intelligence, distributed computing and data mining.
\end{IEEEbiography}

\vspace{11pt}

\vfill

\end{document}